\documentclass[11pt]{article}
\usepackage[preprint]{acl}
\usepackage{times}
\usepackage{latexsym}
\usepackage[T1]{fontenc}
\usepackage[utf8]{inputenc}
\usepackage{microtype}
\usepackage{inconsolata}
\usepackage{graphicx}
\usepackage{amsmath}
\usepackage{amsfonts}
\usepackage{bbm}
\usepackage{booktabs}
\usepackage{multirow}
\usepackage[most]{tcolorbox}

\title{Rubrics as Privileged Information for Open-Ended Generation}

\author{
  Deepika Bablani \quad Ajay Gupta \quad Wanming Chen \\
  Apple \\
  \texttt{deepika\_bablani@apple.com}
}

\begin{document}

\maketitle

\begin{abstract}
On-policy self-distillation (OPSD), where a single model acts as both student and teacher with different contexts, has shown promise in verifiable domains like math, where \emph{hard} privileged information (PI) in the form of ground-truth answers structurally constrains valid continuations. We extend OPSD to open-ended generation using \emph{soft} PI in the form of rubrics that guide preferences but admit many valid responses. Rubrics have served as scalar rewards for reinforcement learning (RL); we show that they provide substantially richer signal as dense PI for distillation, and contrary to intuition, soft rubric PI provides a larger and more effective training signal on student roll-outs than hard reference completion PI in this regime. A reference completion is one point in a set of valid responses, so distilling towards it over-constrains the student, while rubrics specify the preference structure shared across the set of valid responses. We show the effectiveness of using rubrics as PI for open-ended generation across Qwen and Llama model families and show that it outperforms  rubric-as-reward (RaR) RL using HealthBench, a benchmark that grades open-ended health responses against physician-created rubrics, providing dense token-level supervision for open-ended tasks; RuPI beats RaR RL by up to $+0.10$ absolute score and, under matched recipe and KL direction, beats reference-PI by $+0.034$ to $+0.079$ absolute score across three models. We further show that these findings generalize to training on the RubricHub Science corpus and evaluating on ResearchQA: soft rubric PI outperforms both reference-PI distillation and RaR RL ($66.6\%$ vs.\ $64.2\%$ and $57.6\%$).
\end{abstract}

\section{Introduction}
On-policy self-distillation (OPSD), introduced recently in \citep{zhao2026self} and \citep{shenfeld2026self}, demonstrates that a model can effectively serve as its own teacher when conditioned on privileged information (PI), and this can enable learning from self-distillation using on-policy demonstrations. By minimizing the KL divergence between token distributions, the student internalizes what the teacher learned from the PI without needing that information during inference. Existing OPSD methods target \emph{verifiable} domains like math and code and the teacher is conditioned on \emph{hard} PI like ground-truth answers, verified solutions or expert demonstrations that structurally constrain valid continuations. Open-ended generation tasks like long form question answering, medical consultation, creative writing, etc. lack such verifiable PI. For such open-ended generation, the current practice is to either train specialized reward models \citep{ouyang2022training}, use LLM-as-a-judge in the reinforcement learning loop, or use rubrics as rewards for RL. While rubrics as rewards have shown promise, current approaches convert rubrics into a scalar reward (rubric-as-reward; \citealp{gunjal2026rubrics, liu2026openrubrics}), diminishing the richness of the original criteria and the information they encode. We instead propose \textit{\underline{ru}brics as \underline{PI}} (RuPI) and extend OPSD to open-ended generation for non-verifiable tasks using \emph{soft} PI in the form of rubrics. We show that, for non-verifiable tasks, conditioning the teacher on high quality rubrics provides a stronger learning signal than using reference completions.

A per-token KL diagnostic on the untrained base model quantifies this gap: when scored on the student's on-policy roll-outs, rubric conditioning produces $1.7\times$ more per-token KL than reference conditioning, a stronger gradient signal (we develop this mechanism in Section~\ref{sec:diagnostic}). We show empirically that using rubrics as privileged information (RuPI) substantially outperforms both reference-PI distillation and RaR GRPO across two model families on HealthBench, a benchmark that grades open-ended health responses against high-quality, physician-created rubrics, while preserving general capabilities (MMLU, GSM8K, IFEval, TruthfulQA). To show that the findings generalize, we replicate the study on a second domain: training on the Science split of RubricHub \citep{li2026rubrichub}, a large-scale rubric corpus, and evaluating on ResearchQA \citep{yifei2026researchqa}, we show that RuPI outperforms reference PI and that dense distillation outperforms RaR RL (Section~\ref{sec:rubrichub}).

\section{Background}
\subsection{On-policy self-distillation}
OPSD trains a student policy to match a teacher distribution where both are derived from the same base model under different conditioning. The teacher sees PI, whereas the student does not. Classical knowledge distillation \citep{hinton2015distilling,kim2016sequence,sanh2019distilbert} trains a student on fixed sequences generated by a teacher. Instead, on-policy distillation \citep{agarwal2024policy} uses teacher supervision on student generations. Both techniques require a separate, often more capable teacher model. OPSD eliminates the need for a separate teacher while preserving the dense per-token signal of knowledge distillation and the on-policy roll-out structure of OPD/RL. The training objective minimizes the KL divergence between the teacher and student distributions over response tokens generated on-policy from the student: 
\begin{equation}
    \resizebox{\columnwidth}{!}{$\displaystyle
    \mathcal{L} = \mathbb{E}_{x \sim \mathcal{D},\, \hat{y} \sim \pi_S(\cdot|x)} \left[ \sum_{t} D\big(\pi_T(\cdot \mid x, \mathrm{PI}, \hat{y}_{<t}) \;\big\|\; \pi_S(\cdot \mid x, \hat{y}_{<t})\big) \right]
    $},
\end{equation}
where $x$ is a prompt sampled from dataset $\mathcal{D}$, $\hat{y}$ is an on-policy roll-out from the student policy $\pi_S$, $\pi_T$ is the teacher policy conditioned on the same context plus privileged information $\mathrm{PI}$, $\hat{y}_{<t}$ denotes the rollout prefix up to position $t$, and $D$ is a divergence (forward or reverse KL). Throughout, we call $D_{\mathrm{KL}}(\pi_T \,\|\, \pi_S)$ (teacher first) \emph{forward} KL, the mass-covering orientation used by \citet{zhao2026self}, and $D_{\mathrm{KL}}(\pi_S \,\|\, \pi_T)$ (student first) \emph{reverse} KL, the mode-seeking orientation used by \citet{shenfeld2026self} and in our Eq.~\ref{eq:revkl}. 

Prior work \citet{zhao2026self} trains a LoRA \citep{hu2022lora} student that shares base parameters with the teacher, while \citet{shenfeld2026self} full-finetunes with an exponential moving average (EMA) teacher. We study both and find that both improve over base in our setup. The forward-vs-reverse KL choice matters mainly out-of-domain.

\subsection{Hard vs.\ soft privileged information}

The general notion of training a student model with information available to the teacher but not the student dates back to \citet{vapnik2009new} and was unified with distillation by \citet{lopez2015unifying}. In OPSD, the PI takes the form of text injected into the teacher's context. We distinguish PI by how strongly it constrains continuations. \emph{Hard PI} constrains generations structurally: knowing ``the answer is 42'' makes most responses incoherent unless they produce 42, so it cannot be ignored. \emph{Soft PI} biases preferences but admits many valid continuations: ``advise the patient to find a mental health provider'' biases toward including that advice but leaves freedom in style, structure, and content. The distillation signal is correspondingly diffuse under soft PI, spread across many good completions rather than concentrated on one, and we show that this is useful for open-ended tasks.

\subsection{Rubrics as rewards vs.\ rubrics as PI}

Rubric-as-reward methods convert rubrics into a scalar RL reward by taking a weighted sum:
\begin{equation}
    \resizebox{\columnwidth}{!}{$\displaystyle
    R(y) = \frac{\sum_{k=1}^K w_k \cdot \mathbbm{1}[\text{criterion}_k \text{ satisfied by } y]}{\sum_{k=1}^K w_k}
    $},
\end{equation}
where each rubric criterion $k$ carries a weight $w_k$ and $\mathbbm{1}[\cdot]$ is a binary judge satisfaction label. This signal is sparse, one scalar per roll-out. Our approach uses the same rubric criteria as PI, providing dense, token-level supervision through the conditioned teacher's distribution shift; rather than using the per-criterion weights, we group the criteria into ``the response should'' and ``the response should NOT'' lists and include both as natural-language guidance in the teacher's system message. We show that this dense signal substantially outperforms the sparse reward signal in RL.

\subsection{Additional related work}
 \emph{Rubrics as rewards:} \citet{gunjal2026rubrics} (Rubrics as Rewards) and \citet{liu2026openrubrics} convert expert- or synthetically-generated rubrics into a scalar RL reward by aggregating per-criterion satisfaction and optimizing with GRPO; instead, we use the same rubric content as dense privileged information for distillation. \emph{RL post-training:} RLHF \citep{ouyang2022training} and RL from verifiable rewards \citep{guo2025deepseek}, optimized with PPO \citep{schulman2017proximal} or GRPO \citep{shao2024deepseekmath} and combined in open recipes such as T\"ulu~3 \citep{lambert2024tulu}, are the standard post-training tools widely adopted in practice. The EMA-teacher variant we study \citep{shenfeld2026self} is also termed SDPO by \citet{hubotter2026reinforcement}.

\section{Rubrics for OPSD}
\begin{figure}
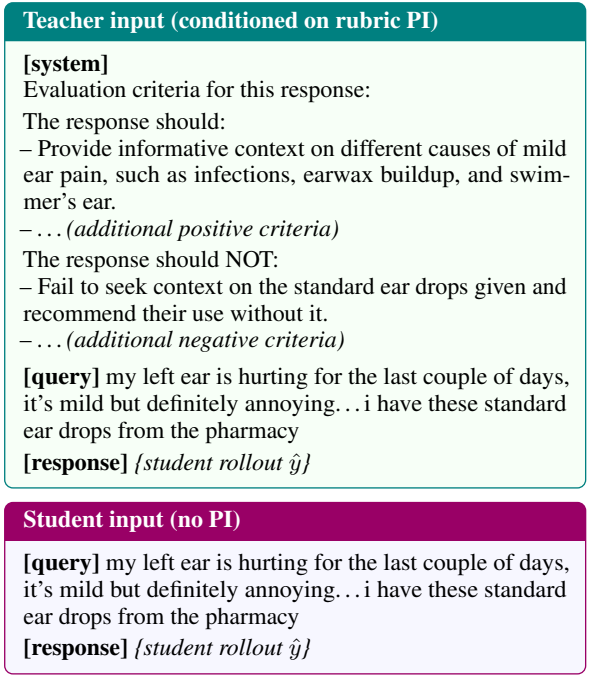

\centering
\small
\begin{tcolorbox}[
    enhanced, colback=green!3, colframe=green!50!blue, boxrule=0.5pt,
    title=\textbf{Teacher input (conditioned on rubric PI)},
    fonttitle=\bfseries\small, left=4pt, right=4pt, top=2pt, bottom=2pt,
]
\textbf{[system]}\\
Evaluation criteria for this response:\\[2pt]
The response should:\\
\textendash\ Provide informative context on different causes of mild ear pain, such as infections, earwax buildup, and swimmer's ear.\\
\textendash\ {\itshape\ldots (additional positive criteria)}\\[2pt]
The response should NOT:\\
\textendash\ Fail to seek context on the standard ear drops given and recommend their use without it.\\
\textendash\ {\itshape\ldots (additional negative criteria)}\\[4pt]
\textbf{[query]} my left ear is hurting for the last couple of days, it's mild but definitely annoying\ldots i have these standard ear drops from the pharmacy\\[2pt]
\textbf{[response]} \emph{\{student rollout $\hat{y}$\}}
\end{tcolorbox}
\begin{tcolorbox}[
    enhanced, colback=blue!3, colframe=blue!40!red, boxrule=0.5pt,
    title=\textbf{Student input (no PI)},
    fonttitle=\bfseries\small, left=4pt, right=4pt, top=2pt, bottom=2pt,
]
\textbf{[query]} my left ear is hurting for the last couple of days, it's mild but definitely annoying\ldots i have these standard ear drops from the pharmacy\\[2pt]
\textbf{[response]} \emph{\{student rollout $\hat{y}$\}}
\end{tcolorbox}
\caption{
Teacher--student conditioning asymmetry on a HealthBench example. The teacher is given the rubric as a system message before the query; the student sees only the query. Both contexts include the \emph{same} on-policy rollout $\hat{y}$ generated by the student; the teacher scores it under its rubric-conditioned distribution rather than producing its own response. For the reference-PI baseline, the system message is replaced with \texttt{"Reference response:\textbackslash n\{ideal completion\}"}.
}
\label{fig:prompts}
\end{figure}
We train the student with full fine-tuning, using an EMA of the student parameters for the teacher: $\theta_T \leftarrow \mu \theta_S + (1-\mu)\theta_T$ after each step ($\mu=0.02$). Given a prompt $x$, rubric $r$, and the student's on-policy roll-out $\hat{y} \sim \pi_S(\cdot|x)$, the teacher's response is additionally conditioned on the rubric: $\pi_T(\cdot \mid x, r, \hat{y}_{<t})$, whereas the student has no rubric access: $\pi_{S,\theta}(\cdot \mid x, \hat{y}_{<t})$ (Figure~\ref{fig:prompts} shows a real example). We use reverse KL with per-token clipping:
\begin{equation}
    \resizebox{\columnwidth}{!}{$\displaystyle
    \mathcal{L}(x, r, \hat{y}) = \frac{1}{|\hat{y}|} \sum_t \min\!\Big(\tau,\; D_{\mathrm{KL}}\big(\pi_{S,\theta}(\cdot \mid x, \hat{y}_{<t}) \;\big\|\; \pi_T(\cdot \mid x, r, \hat{y}_{<t})\big)\Big)
    $},
    \label{eq:revkl}
\end{equation}

Reverse KL induces mode-seeking behavior, encouraging the student to focus on high-density regions of the teacher distribution instead of covering its full support. This keeps the student close to the base distribution while selectively adopting rubric-aligned behaviors, i.e., it enables targeted behavioral transfer while minimizing drift from the base policy, and we show that this helps in preserving the student's instruction-following capability.

Our LoRA variant uses adapters ($r{=}64$, $\alpha{=}128$, all linear layers) with a frozen base-model teacher; the rank constraint provides implicit regularization. \citet{zhao2026self} originally use forward KL; we run the LoRA arm in both KL directions and find forward and reverse comparable in-domain, but recommend reverse KL because it better preserves out-of-domain capabilities (Section~\ref{sec:cross_domain_new}).

We use per-token KL clipping at $\tau=5.0$ and generate on-policy roll-outs from the student. The KL divergence at each response token is computed over the model's full vocabulary distribution rather than a top-$K$ approximation, so each roll-out produces a dense supervision signal across every response position. The hyperparameters $r{=}64$, $\alpha{=}128$, EMA decay $\mu{=}0.02$, and per-token KL clip $\tau=5.0$ are inherited from \citet{zhao2026self} and \citet{shenfeld2026self}; we do not re-tune them.

\section{Experiments}
\subsection{Experimental Setup}

\paragraph{Evaluation} We evaluate our method on HealthBench \citep{arora2025healthbench}, a benchmark designed to measure the capabilities of AI systems for health. HealthBench includes 5,000 realistic health conversations, each with a custom physician-created rubric to grade model responses. Each rubric specifies a set of weighted criteria (positive criteria the response should satisfy and negative criteria it should avoid), which we use as the rubric privileged information $r$ during training. 

We use 2000/500/2500 train/validation/test splits and report HealthBench rubric satisfaction following the official protocol.\footnote{Train and test have closely matching theme proportions (largest difference $0.03$) and rubric sizes (mean $11.4$ criteria).} We also evaluate on the official 1{,}000-prompt HealthBench Hard subset (the most challenging slice, disjoint from our training pool). On the Hard subset this published score clips to $0$ for almost every method at the open-source 7B/14B scale, so there we additionally report HealthBench's non-clipped \emph{cluster}-level score, which aggregates per-rubric-cluster satisfaction and remains informative across methods at our scale. All reported scores are bootstrap means, with standard errors $\le 0.008$ on test ($n{=}2{,}500$) and $\le 0.020$ on val ($n{=}500$), so test differences below $\sim 0.01$ should be treated as within noise.\footnote{We verified that response length does not drive the method ranking on HealthBench: per-example Pearson correlations between response character length and rubric satisfaction score are small in magnitude ($|r| \le 0.08$) across all methods, and per-method mean response lengths span only $1{,}157$--$1{,}700$ characters on the Qwen 7B test set. A post-hoc length-adjusted score preserves the ordering reported in the tables below.}

\paragraph{Models and training} We use Qwen2.5-7B-Instruct \citep{qwen2025qwen25technicalreport} as our base model and additionally report scale-up results with Qwen2.5-14B-Instruct. We also evaluate generalization to Llama-3.1-8B-Instruct \citep{grattafiori2024llama}. All distillation runs train for 200 steps on 2{,}000 HealthBench prompts with an effective batch size of 32 (32 unique prompts in a batch with one on-policy roll-out per prompt). For LoRA-based runs we attach low-rank adapters with $r{=}64$ and $\alpha{=}128$, trained at a learning rate of $1\mathrm{e}{-5}$. For full fine-tuning, we use an EMA teacher (decay rate $\mu{=}0.02$) at a lower learning rate of $5\mathrm{e}{-6}$ to compensate for the larger number of trainable parameters. We compare against a RaR GRPO \citep{shao2024deepseekmath} baseline that also sees 32 unique prompts per batch, with $G{=}8$ completions per prompt for a total batch size of 256. The reward signal is provided by GPT-4.1 as the external LLM judge that evaluates each generation against the rubric criteria, returning a per-criterion binary satisfaction label that is aggregated into a scalar reward via the rubric's point weights. We use the same judge (GPT-4.1) for the GRPO reward as for evaluation, so the RL baseline is not disadvantaged by any policy-judge mismatch. Note that GRPO consumes $8{\times}$ as many on-policy generations per training prompt as RuPI ($G{=}8$ vs.\ a single rollout). Thus, RuPI uses $8{\times}$ fewer sampled rollouts per prompt while outperforming GRPO at matched prompt count. Each configuration was trained once; reported test scores are bootstrap means over evaluation examples from the validation-selected checkpoint.

\paragraph{Cross-domain evaluation.} We additionally evaluate each checkpoint on a four-task cross-domain suite using \texttt{lm-evaluation-harness} \citep{eval-harness} with default settings: MMLU \citep{hendrycks2020measuring} (5-shot, world knowledge across 57 subjects), GSM8K \citep{cobbe2021training} (8-shot CoT, grade-school math reasoning), IFEval \citep{zhou2023instruction} (zero-shot, verifiable instruction-following constraints), and TruthfulQA \citep{lin2022truthfulqa} (zero-shot, factual accuracy on adversarial questions) to check that gains on HealthBench do not come at the cost of loss of other general capabilities. MMLU, GSM8K, and TruthfulQA together probe whether distillation degrades the model's underlying knowledge, reasoning, or factual calibration, while IFEval specifically tests whether teacher-style transfer has shifted the model away from following user instructions.

\subsection{Rubrics provide trainable signal where reference PI does not}\label{sec:diagnostic}

Before training, we test a precondition for distillation to work: does conditioning the teacher on a soft rubric actually shift its output distribution relative to the unconditioned student? If the two distributions are nearly identical, the training signal at the start of training is weak. A rubric specifies abstract criteria rather than a target response, and this may not lead to a substantially different next-token distribution from the unconditioned generation.

We measure this on the untrained base model. Using Qwen2.5-7B-Instruct as both teacher and student, we compute per-token KL divergence between the rubric-conditioned teacher and the unconditioned student over 213 held-out HealthBench prompts, averaged across the response tokens of an on-policy roll-out sampled from the student. The student's natural roll-out is what the student will be trained on, and the per-token KL on it is the gradient signal that the student will actually see during training. We report two variants: \emph{Mean KL} over all response tokens and \emph{Body KL}, which restricts the average to the middle 80\% of each response (positions $[0.1n, 0.9n]$). The body restriction excludes opening boilerplate (greetings, ``Sure, here's\ldots'') and closing boilerplate (sign-offs, end-of-sequence cues) at the response boundaries, where conditioning has little effect and the KL is dominated by stylistic regularities rather than content. We repeat the same for reference PI.

\begin{table}[h]
\centering
\caption{Per-token teacher-student KL on the base model (Qwen2.5-7B-Instruct), measured on roll-outs sampled from the student. Rubric PI produces $1.7\times$ \emph{more} per-token KL than reference PI.}
\vspace{0.5em}
\small
\begin{tabular}{lcc}
\toprule
PI type & Mean KL & Body KL \\
\midrule
Physician rubrics & \textbf{0.373} & \textbf{0.326} \\
Reference completions & 0.212 & 0.189 \\
\bottomrule
\end{tabular}
\label{tab:diagnostic}
\end{table}

Both PI types produce substantial asymmetry, confirming the precondition: rubric conditioning shifts the teacher's distribution by a meaningful amount even though the criteria never specify a target response. Table~\ref{tab:diagnostic} shows that rubric PI provides a stronger training signal in the form of per-token KL on the student's rollout than reference PI.

To understand \emph{why} reference PI is less useful for training, we measure how the teacher's probability mass is distributed across the valid response set. For each held-out HealthBench prompt we score the teacher's per-token negative log-likelihood (NLL) on the gold completion and on a diverse set of other rubric-satisfying responses ($423$ held-out HealthBench prompts, $3{,}352$ GPT-4.1-generated candidates; full protocol in Appendix~\ref{app:concentration}). As Table~\ref{tab:concentration} shows, under hard (reference) PI the gold is $1.18$ nats \emph{more} probable per token than other valid responses, whereas under soft (rubric) PI it is $0.70$ nats \emph{less} probable, the gold is thus $\exp(1.88)\approx 6.5\times$ more over-concentrated per token under hard PI. Reference PI piles mass on one arbitrary point in the valid set; rubric PI spreads it across the set the student is exploring.

\begin{table}[h]
\centering
\caption{Teacher per-token NLL under each PI conditioning (Qwen2.5-7B-Instruct teacher). Hard PI concentrates $6.5\times$ more probability mass \emph{per token} on the gold (relative to other valid responses) than soft PI; robust to the candidate generator ($7.1\times$ with an independent Qwen2.5-14B + rubric-LoRA generator).}
\vspace{0.5em}
\small
\begin{tabular}{lcc}
\toprule
per-token NLL & Rubric PI & Reference PI \\
\midrule
gold completion & 2.32 & \textbf{0.38} \\
other good responses & 1.62 & 1.56 \\
\midrule
\shortstack[l]{over-concentration\\(other-good $-$ gold, nats)} & $-0.70$ & $+1.18$ \\
\bottomrule
\end{tabular}
\label{tab:concentration}
\end{table}

To make explicit how the signal depends on \emph{where} the KL is scored, we re-run the per-token KL measurement of Table~\ref{tab:diagnostic} with one variable changed: we score on the held-out gold completion instead of the student's on-policy rollout. Everything else is identical. Rubric PI induces similar KL on student roll-outs and on gold completions ($0.326$ vs.\ $0.313$), whereas reference PI induces far larger KL on the gold than on student roll-outs ($1.666$ vs.\ $0.189$, an $8.8\times$ gap) (Table~\ref{tab:kl_response}). Since training optimizes KL on the student's own trajectories, RuPI is the stronger learning signal, even though reference PI looks more powerful when scored at the gold.

\begin{table}[h]
\centering
\caption{Per-token \emph{Body} KL (middle 80\% of each response) between PI-conditioned teacher distributions on Qwen2.5-7B-Instruct, scored on either the student's on-policy roll-out or the held-out gold completion. Rubric KL is invariant to the response choice; reference KL is $8.8\times$ larger on the gold than on a student roll-out.}
\vspace{0.5em}
\small
\begin{tabular}{lcc}
\toprule
PI type & Student rollout & Gold completion \\
\midrule
Rubric & $0.326$ & $0.313$ \\
Reference & $0.189$ & $\mathbf{1.666}$ \\
\bottomrule
\end{tabular}
\label{tab:kl_response}
\end{table}

\section{Main Results}\label{sec:main}

\subsection{HealthBench: in-domain performance}\label{sec:hb_new}

Tables~\ref{tab:hb_7b_new}, \ref{tab:hb_14b_new}, and~\ref{tab:hb_llama_new} report each method at its best-by-val checkpoint, evaluated on the 2{,}500-prompt held-out test split, and Figure~\ref{fig:trajectories} (Appendix~\ref{app:trajectories}) shows the per-step validation trajectories.

\begin{table}[t]
\centering
\caption{Qwen2.5-7B-Instruct on HealthBench. Best (val-selected) checkpoint per method. ``ckpt'' is the val-selected step. Standard errors $\le 0.008$ on test (n${=}2500$). Fwd/Rev: forward/reverse KL; FT: full fine-tuning.}
\vspace{0.5em}
\footnotesize
\begin{tabular}{lllccc}
\toprule
Method & KL & PI & ckpt & Val & Test \\
\midrule
Base & --- & --- & --- & 0.147 & 0.169 \\
GRPO & --- & \shortstack[l]{rubric\\(reward)} & 50 & 0.178 & 0.193 \\
\midrule
LoRA & Fwd & rubric & 200 & \textbf{0.223} & \textbf{0.215} \\
LoRA & Rev & rubric & 200 & 0.213 & 0.203 \\
FT & Fwd & rubric & 100 & 0.207 & 0.212 \\
FT & Rev & rubric & 200 & 0.197 & 0.205 \\
\midrule
LoRA & Fwd & reference & 150 & 0.169 & 0.172 \\
LoRA & Rev & reference & 150 & 0.164 & 0.169 \\
FT & Fwd & reference & 200 & 0.168 & 0.165 \\
FT & Rev & reference & 200 & 0.151 & 0.159 \\
\midrule
SFT & --- & \shortstack[l]{rubric\\(RCG)} & 150 & 0.184 & 0.169 \\
SFT & --- & reference & 200 & 0.140 & 0.136 \\
\bottomrule
\end{tabular}
\label{tab:hb_7b_new}
\end{table}

\begin{table}[t]
\centering
\caption{Qwen2.5-14B-Instruct on HealthBench. Best (val-selected) checkpoint per method. Fwd/Rev: forward/reverse KL; FT: full fine-tuning.}
\vspace{0.5em}
\footnotesize
\begin{tabular}{lllccc}
\toprule
Method & KL & PI & ckpt & Val & Test \\
\midrule
Base & --- & --- & --- & 0.247 & 0.269 \\
GRPO & --- & \shortstack[l]{rubric\\(reward)} & 50 & 0.264 & 0.268 \\
\midrule
LoRA & Fwd & rubric & 200 & 0.337 & 0.324 \\
LoRA & Rev & rubric & 150 & 0.334 & 0.325 \\
FT & Fwd & rubric & 200 & \textbf{0.363} & \textbf{0.347} \\
FT & Rev & rubric & 150 & 0.330 & 0.339 \\
\midrule
LoRA & Fwd & reference & 200 & 0.287 & 0.274 \\
LoRA & Rev & reference & 100 & 0.275 & 0.267 \\
FT & Fwd & reference & 100 & 0.273 & 0.267 \\
FT & Rev & reference & 150 & 0.274 & 0.258 \\
\midrule
SFT & --- & \shortstack[l]{rubric\\(RCG)} & 250 & 0.236 & 0.226 \\
SFT & --- & reference & 100 & 0.253 & 0.251 \\
\bottomrule
\end{tabular}
\label{tab:hb_14b_new}
\end{table}

\begin{table}[t]
\centering
\caption{Llama-3.1-8B-Instruct on HealthBench. Best (val-selected) checkpoint per method. Fwd/Rev: forward/reverse KL; FT: full fine-tuning.}
\vspace{0.5em}
\footnotesize
\begin{tabular}{lllccc}
\toprule
Method & KL & PI & ckpt & Val & Test \\
\midrule
Base & --- & --- & --- & 0.114 & 0.141 \\
GRPO & --- & \shortstack[l]{rubric\\(reward)} & 50 & 0.112 & 0.116 \\
\midrule
LoRA & Fwd & rubric & 150 & 0.169 & 0.197 \\
LoRA & Rev & rubric & 150 & \textbf{0.206} & 0.204 \\
FT & Fwd & rubric & 200 & 0.185 & 0.196 \\
FT & Rev & rubric & 200 & 0.204 & \textbf{0.213} \\
\midrule
LoRA & Fwd & reference & 150 & 0.114 & 0.124 \\
LoRA & Rev & reference & 100 & 0.121 & 0.125 \\
FT & Fwd & reference & 50 & 0.108 & 0.104 \\
FT & Rev & reference & 50 & 0.135 & 0.134 \\
\midrule
SFT & --- & \shortstack[l]{rubric\\(RCG)} & 300 & 0.169 & 0.176 \\
SFT & --- & reference & 300 & 0.196 & 0.189 \\
\bottomrule
\end{tabular}
\label{tab:hb_llama_new}
\end{table}

\paragraph{RuPI outperforms RaR GRPO across all three base models.} On the held-out test split, all four RuPI configurations beat GRPO on every base (Tables~\ref{tab:hb_7b_new}--\ref{tab:hb_llama_new}) while using $8\times$ fewer on-policy generations per training prompt and no external judge. Notably, GRPO optimizes \emph{directly} against the same GPT-4.1 rubric judge used to score it, yet is still beaten by RuPI, which never queries a judge during training---dense rubric distillation extracts more from a rubric than optimizing it as a scalar reward. The margin is largest on the weakest base (Llama: RuPI up to $0.213$ vs.\ GRPO $0.116$, $+0.10$) and narrows on Qwen 7B ($+0.01$ to $+0.02$), where the GPT-4.1-reward GRPO is a strong baseline. GRPO makes only a small gain over base on Qwen~7B ($0.169{\to}0.193$), is essentially flat on Qwen~14B ($0.269{\to}0.268$), and slightly declines on Llama ($0.141{\to}0.116$): its trajectory-level reward is too sparse to match dense distillation.

\paragraph{Soft (rubric) PI outperforms hard (reference) PI consistently.} Holding the recipe fixed (reverse KL), the best RuPI variant beats the best reference-PI variant on every base, by $+0.034$ (Qwen 7B) up to $+0.079$ (Llama 8B). This empirically realizes the diagnostic of Section~\ref{sec:diagnostic}: rubric PI produces a stronger, better-aimed KL signal on the student's own roll-outs. The ordering also holds under forward KL (e.g., 14B full fine-tuning: RuPI $0.347$ vs.\ reference $0.267$; 7B: $0.212$ vs.\ $0.165$).

\paragraph{Forward and reverse KL are comparable in-domain.} Across recipes and scales the two directions land within $\sim 0.015$ of each other in-domain; forward KL holds a slight edge for rubric variants on Qwen in most settings (LoRA $0.215$ vs.\ $0.203$ and full fine-tuning $0.212$ vs.\ $0.205$ at 7B; full fine-tuning $0.347$ vs.\ $0.339$ at 14B), with the 14B LoRA pair effectively tied ($0.324$ vs.\ $0.325$). The choice has modest in-domain effect but matters out-of-domain, where we recommend reverse KL for capability preservation (Section~\ref{sec:cross_domain_new}).

\paragraph{SFT baselines: information vs.\ mechanism.} To separate the rubric \emph{information} from the on-policy dense-KL \emph{mechanism}, we add two supervised fine-tuning baselines (Tables~\ref{tab:hb_7b_new}--\ref{tab:hb_llama_new},  bottom). \emph{Reference-SFT} fine-tunes on reference completions, the same target as reference PI, and \emph{RCG-SFT} (rubric-conditioned-generation SFT) fine-tunes on the base model's own responses generated with the rubric in context, with the rubric stripped at training time, so it derives its targets from the same rubric information that RuPI uses, injected offline as fixed targets.\footnote{Both are learning-rate swept over $\{5\mathrm{e}{-6}, 1\mathrm{e}{-5}, 2\mathrm{e}{-5}, 3\mathrm{e}{-5}\}$ with best-by-val checkpoint selection.} Neither recovers RuPI's gain, i.e., matched rubric information delivered by offline SFT does not reproduce the rubric-distillation gain: the on-policy dense-KL mechanism, along with the rubric content, is what makes RuPI work. Out of domain (Table~\ref{tab:cd_7b_new}, \ref{tab:cd_14b_new}, and~\ref{tab:cd_llama_new}) and HealthBench Hard results (Table~\ref{tab:sft_hard}) are in Appendix~\ref{app:cross_domain_full} and Appendix~\ref{app:sft_hard}, and further demonstrate that SFT performs worse than RuPI while causing significant regression in cross-domain evaluation.

\subsection{HealthBench Hard}\label{sec:hard_new}

We additionally evaluate each method's best checkpoint on the official HealthBench Hard subset. At the open-source-7B-and-8B scale, the published rubric-satisfaction score collapses to $0$ for almost every method, so we report HealthBench's non-clipped facet scores. The two facets that drive our conclusions, the cluster-level aggregate (Cluster) and the accuracy axis (Acc), are in Table~\ref{tab:hard_summary}, reporting the strongest reference-PI and RuPI variant per metric. The remaining axes (communication quality, instruction following, completeness, context awareness) saturate near-uniformly across methods or clip to zero and are omitted.

\begin{table}[t]
\centering
\caption{HealthBench Hard. We report the cluster-level aggregate (Cluster) and the accuracy axis (Acc). RefPI / RuPI is the strongest reference-PI (RefPI) / rubric-PI (RuPI) variant per metric; \textbf{bold} is the best per row.}
\vspace{0.5em}
\footnotesize
\begin{tabular}{llcccc}
\toprule
Model & Metric & Base & GRPO & RefPI & RuPI \\
\midrule
\multirow{2}{*}{Qwen 7B} & Cluster & 0.503 & 0.535 & 0.501 & \textbf{0.542} \\
 & Acc & \textbf{0.043} & 0.008 & 0.015 & 0.038 \\
\midrule
\multirow{2}{*}{Qwen 14B} & Cluster & 0.698 & 0.706 & 0.701 & \textbf{0.778} \\
 & Acc & 0.106 & 0.105 & 0.101 & \textbf{0.156} \\
\midrule
\multirow{2}{*}{Llama 8B} & Cluster & 0.532 & 0.515 & 0.545 & \textbf{0.625} \\
 & Acc & 0.000 & 0.000 & 0.000 & \textbf{0.077} \\
\bottomrule
\end{tabular}
\label{tab:hard_summary}
\end{table}

The strongest separation on HealthBench Hard is on the accuracy axis, where RuPI consistently exceeds both GRPO and reference-PI distillation (Table~\ref{tab:hard_summary}). The effect is most pronounced on Llama 8B, where \emph{only} RuPI variants achieve non-zero accuracy. RuPI also leads on the cluster-level aggregate for all three models.

\subsection{Cross-domain capabilities}\label{sec:cross_domain_new}

To check that in-domain gains do not come at the cost of general capability, we evaluate each method's best (val-selected) checkpoint on a four-task cross-domain suite (MMLU, GSM8K, IFEval, TruthfulQA). Three of the four tasks barely move across methods; the only axis that separates methods is IFEval, which we report as the change relative to base ($\Delta$IFEval) in Table~\ref{tab:cd_ifeval}, with the full four-task tables in Appendix~\ref{app:cross_domain_full}.

\begin{table}[t]
\centering
\caption{Change in IFEval relative to base ($\Delta$IFEval). Forward KL degrades instruction-following on Qwen; reverse KL stays near base; on Llama forward-KL FT leaves IFEval roughly flat. The Qwen-14B base scores below Qwen-7B on IFEval, a genuine instruction-following inversion between these instruct checkpoints, not a parsing artifact, so we compare only within-model $\Delta$. Full four-task tables (MMLU, GSM8K, IFEval, TruthfulQA) in Appendix~\ref{app:cross_domain_full}. Fwd/Rev: forward/reverse KL; FT: full fine-tuning; ref: reference.}
\vspace{0.5em}
\small
\begin{tabular}{lccc}
\toprule
Method & Qwen 7B & Qwen 14B & Llama 8B \\
\midrule
GRPO (rubric) & $+0.013$ & $-0.004$ & $-0.015$ \\
\midrule
LoRA-Fwd rubric & $-0.046$ & $-0.041$ & $+0.014$ \\
LoRA-Rev rubric & $+0.021$ & $-0.032$ & $-0.012$ \\
FT-Fwd rubric & $-0.038$ & $-0.034$ & $-0.000$ \\
FT-Rev rubric & $-0.007$ & $-0.010$ & $-0.010$ \\
\midrule
LoRA-Fwd ref & $-0.009$ & $-0.019$ & $-0.006$ \\
LoRA-Rev ref & $+0.019$ & $-0.019$ & $+0.005$ \\
FT-Fwd ref & $-0.011$ & $-0.006$ & $+0.009$ \\
FT-Rev ref & $+0.021$ & $-0.012$ & $-0.012$ \\
\bottomrule
\end{tabular}
\label{tab:cd_ifeval}
\end{table}

\paragraph{Forward KL degrades IFEval on Qwen; on Llama both directions stay near base.} Forward-KL distillation lowers IFEval relative to reverse KL on both Qwen scales: the largest drops are forward+rubric ($-0.046$ on 7B, $-0.041$ on 14B), with forward+reference also regressing on 14B ($-0.019$). Reverse-KL variants stay at or above base on Qwen 7B (rubric $+0.021$, reference $+0.019$) and within $0.032$ of base on 14B ($-0.032$ to $-0.012$). On Llama, IFEval moves little under \emph{both} KL directions ($-0.012$ to $+0.014$), and forward KL does not degrade it the way it does on Qwen. The Qwen out-of-distribution behaviour is the main reason we recommend reverse KL despite forward KL's slight in-domain edge (Section~\ref{sec:hb_new}). We observe no material degradation on MMLU, GSM8K, or TruthfulQA (full numbers in Appendix~\ref{app:cross_domain_full}).

\subsection{On-policy roll-outs are critical}\label{sec:onpolicy}

All RuPI results above train on the student's \emph{own} on-policy roll-outs ($\lambda{=}1$ in the GKD \citep{agarwal2024policy} sense: every training sequence is sampled from the current student). To test whether the rubric advantage depends on this choice, we re-run the full Qwen2.5-7B panel at $\lambda{=}0$, i.e., pure off-policy distillation, where the student is trained on roll-outs generated by the \emph{teacher} rather than itself. We run $\lambda{=}0$ in both teacher regimes, a frozen cached teacher and an EMA teacher with regenerated targets, and find the two within noise of each other; we report the better value per cell here and defer the full $12$-run panel to Appendix~\ref{app:lambda_panel}.

\begin{table}[h]
\centering
\caption{On-policy ($\lambda{=}1$) vs.\ off-policy ($\lambda{=}0$) distillation on Qwen2.5-7B-Instruct (HealthBench test, $n{=}2500$; base $=0.169$). On-policy values are the best rubric/reference variant from Table~\ref{tab:hb_7b_new}; off-policy values are the best of the cached/EMA $\lambda{=}0$ panel.}
\vspace{0.5em}
\small
\begin{tabular}{lccc}
\toprule
Roll-out source & rubric PI & reference PI \\
\midrule
On-policy ($\lambda{=}1$)  & \textbf{0.215} & 0.172  \\
Mixed ($\lambda{=}0.5$)    & 0.178 & --- \\
Off-policy ($\lambda{=}0$) & 0.154 & 0.156  \\
\midrule
$\Delta$ (on $-$ off) & $+0.061$ & $+0.016$ \\
\bottomrule
\end{tabular}
\label{tab:onpolicy}
\end{table}

Every $\lambda{=}0$ variant lands at or below the base model (best off-policy $0.156$ vs.\ base $0.169$), and the rubric advantage collapses: rubric and reference PI are indistinguishable off-policy ($0.154$ vs.\ $0.156$, within the $\le 0.008$ test standard error), versus a $+0.043$ rubric margin on-policy. We also run the intermediate $\lambda{=}0.5$ mix; holding the recipe fixed (LoRA, forward KL), rubric quality rises \emph{monotonically} with the on-policy fraction ($0.154 \to 0.167 \to 0.215$ for $\lambda{=}0,0.5,1$). Off-policy distillation loses the rubric edge and fails to improve over the base, and the same flatness holds on HealthBench Hard (all $\lambda{=}0$ cluster scores $0.47$--$0.52$, near the base $0.503$). This is in line with the mechanism in Section~\ref{sec:diagnostic}: off-policy training scores the teacher on its own roll-outs rather than the student's, removing the on-policy coverage that gives rubric PI its edge, so the rubric advantage is a property of rubrics \emph{combined with on-policy roll-outs}, not of rubrics alone.

\subsection{Second domain: RubricHub}\label{sec:rubrichub}

Our HealthBench results are confined to a single domain (open ended medical Q\&A). To test that the ordering, rubric PI $>$ reference PI (H1) and dense distillation $>$ RaR RL (H2), generalizes, we replicate the study on RubricHub \citep{li2026rubrichub}, a recent large-scale rubric corpus of $\sim$110K query--rubric pairs. We focus on RubricHub's Science domain, evaluated with ResearchQA \citep{yifei2026researchqa}, whose rubric-coverage metric is an open-ended analogue of HealthBench's rubric grading. 

\paragraph{Setup} RubricHub ships two aligned datasets: a rubric-conditioned RL set (queries with per-domain rubrics) and a rejection-sampling SFT set (queries with best-of-six GPT-5.1 answers). We take their \emph{intersection} on query text, so every training prompt carries all three signals we need: the \textit{rubric} (soft PI for RuPI), a \textit{best-of-six reference answer} (reference PI), and the rubric criteria as the \textit{reward} for the GRPO baseline. We train on 1{,}000 Science prompts on Qwen2.5-7B-Instruct, with recipes identical to HealthBench (LoRA in both KL directions, full fine-tuning with an EMA teacher; GRPO with the stabilized $\beta{=}0.1$ recipe). We use ResearchQA's official protocol: mean rubric-criterion coverage over its 3{,}750-question held-out test split with a \texttt{gpt-4.1-mini} judge, and its 703-question \texttt{valid} split for best-by-val checkpoint selection (sweep over steps 50/100/150/200).

\begin{table}[t]
\centering
\caption{RubricHub Science on ResearchQA, Qwen2.5-7B-Instruct. Best (val-selected) checkpoint per method}
\vspace{0.5em}
\footnotesize
\begin{tabular}{lllccc}
\toprule
Method & KL & PI & ckpt & Val & Test \\
\midrule
Base & --- & --- & --- & --- & 0.574 \\
GRPO & --- & \shortstack[l]{rubric} & 200 & 0.570 & 0.576 \\
\midrule
LoRA & Fwd & rubric & 200 & 0.658 & 0.660 \\
LoRA & Rev & rubric & 150 & 0.649 & 0.647 \\
FT & Fwd & rubric & 200 & \textbf{0.670} & \textbf{0.666} \\
FT & Rev & rubric & 50 & 0.650 & 0.644 \\
\midrule
LoRA & Fwd & reference & 100 & 0.632 & 0.632 \\
LoRA & Rev & reference & 100 & 0.627 & 0.629 \\
FT & Fwd & reference & 200 & 0.639 & 0.642 \\
FT & Rev & reference & 200 & 0.630 & 0.634 \\
\bottomrule
\end{tabular}
\label{tab:rubrichub_7b}
\end{table}

The HealthBench ordering replicates cleanly here (Table \ref{tab:rubrichub_7b}). GRPO does not improve over base, echoing that RaR RL is a weak signal (H2). Soft-PI distillation, by contrast, moves ResearchQA substantially: the best RuPI variant scores $+0.092$ over base and $+0.090$ over GRPO. RuPI also beats reference PI under every matched recipe, realizing H1 in a second domain: rubrics, which characterize the whole set of good answers, transfer better than a single reference answer. Val and test track closely throughout, the ordering is decided on the \texttt{val} split and confirmed independently on test. The precondition diagnostic of Section~\ref{sec:diagnostic}, i.e.\ per-token teacher-student Body KL under rubric vs.\ reference PI on the student's roll-outs, and the mass-concentration measurement are in Appendix~\ref{app:rubrichub_kl}.

\section{Conclusion}
We show that on-policy self-distillation extends to open-ended generation when rubrics serve as PI for the teacher. The central finding is that soft rubric PI provides a stronger and better-aimed training signal on the student's roll-outs than hard reference completion PI. Reference conditioning sharpens the teacher's mass at the gold, but the gold is one point in a large valid response space and the student's roll-outs are far from it, so the conditioning effect on the student's actual trajectories is weak. Soft rubric criteria specify preferences over the whole valid response space the student is exploring, so the same student roll-outs move under conditioning in ways that meaningfully match the rubric. Across three models, RuPI beats RaR GRPO on HealthBench (by up to $+0.10$), with $8\times$ fewer on-policy generations per training prompt and no external judge, and the same pattern replicates on the held-out HealthBench Hard split. The findings also hold in a second domain: trained on RubricHub Science and evaluated on ResearchQA, RuPI outperforms both reference-PI distillation and RaR RL. Together, these results suggest that, for non-verifiable tasks where evaluation rubrics exist, rubrics are better used as dense PI for distillation than as sparse rewards for RL.

\section*{Limitations}
(i) We evaluate on two open-ended domains, medical Q\&A (HealthBench) and scientific Q\&A (RubricHub Science, via ResearchQA), with the HealthBench study spanning two model families (Qwen2.5, Llama-3.1) and three sizes (Llama 8B, Qwen 7B, Qwen 14B). Both are rubric-coverage-graded question-answering tasks; extending the recipe to structurally different non-verifiable tasks, e.g., long-form creative writing or multi-turn assistant dialogue, is a natural next step. (ii) LoRA hyperparameters ($r{=}64$, $\alpha{=}128$), EMA decay ($\mu{=}0.02$), and the KL clip threshold ($\tau{=}5.0$) are inherited from prior work; these values worked well and so we did not do a hyperparameter sensitivity sweep. (iii) Our cross-domain suite (MMLU, GSM8K, IFEval, TruthfulQA) covers knowledge, reasoning, instruction-following, and factual calibration; contamination-resistant instruction-following benchmarks such as IFBench \citep{pyatkin2026generalizing} would further strengthen the OOD claim. (iv) Our study assumes the rubric and the reference answer are both \emph{high-quality} and \emph{aligned with the evaluation criteria}: the premise of RuPI is that a good rubric characterizes the whole distribution of good answers, whereas a reference supplies only one such answer. When rubrics are low-quality, synthetically generated, or \emph{misaligned with what the downstream task actually grades}, so that the rubric and the reference point at different targets, the ordering we report (rubric PI $>$ reference PI) need not hold, and rubric conditioning may even transfer criteria that the evaluation does not reward. Characterizing which rubrics help versus hurt, and automatically curating or repairing rubrics to meet this alignment precondition, is beyond our scope and left to future work.

\section*{Ethical Considerations}
Our experiments include medical question-answering benchmarks, but the resulting models are research prototypes and should not be used for diagnosis, treatment, or clinical decision-making. RuPI transfers behaviors specified by rubrics; inaccurate, incomplete, biased, or malicious rubrics may therefore transfer harmful guidance or omissions. This risk is particularly important in high-stakes domains such as health. Any deployment would require domain-expert validation of the rubrics, safety evaluations beyond rubric satisfaction, monitoring for failures, and appropriate human oversight. Training multiple 7B–14B parameter models and using external LLM judges also incurs computational and environmental costs. RuPI uses fewer sampled rollouts and does not require an external judge during training relative to our GRPO baseline, but we do not provide a complete energy or carbon accounting.
\bibliography{references}

\appendix

\section{Mass-concentration measurement protocol}\label{app:concentration}

This appendix details the per-token NLL measurement summarized in Section~\ref{sec:diagnostic} (Table~\ref{tab:concentration}). For each held-out HealthBench prompt, we sample $8$ candidate responses from GPT-4.1 at temperature $0.9$ to form a diverse set of rubric-satisfying answers ($423$ prompts with valid gold completions yield $3{,}352$ candidates). For each candidate and the held-out gold completion, we compute the teacher's per-token negative log-likelihood under hard PI (system message $=$ \texttt{"Reference response:\textbackslash n\{gold\}"}) and soft PI (system message $=$ rubric criteria text). The teacher is the same Qwen2.5-7B-Instruct base model used elsewhere. The robustness check ($7.1\times$) repeats the measurement with candidates drawn from an independent Qwen2.5-14B + rubric-LoRA generator on the same prompts.

\section{Validation trajectories}\label{app:trajectories}

\begin{figure*}[t]
\centering
\includegraphics[width=\linewidth]{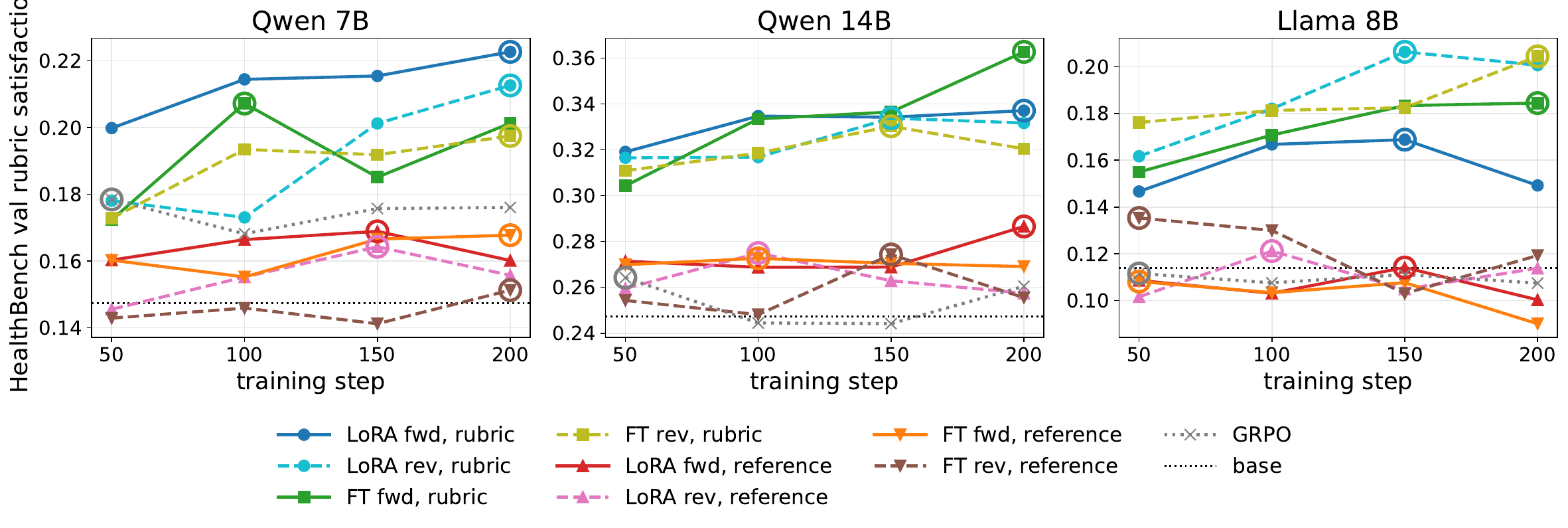}
\caption{Per-step HealthBench \emph{validation} trajectories (GPT-4.1 judge, $T{=}1.0$) for each base model. Each line is a (recipe, PI, KL-direction) combination; the circled marker is that line's best-by-val checkpoint, whose held-out \emph{test} score is the value reported in Tables~\ref{tab:hb_7b_new}--\ref{tab:hb_llama_new}. The dashed black line is the base model.}
\label{fig:trajectories}
\end{figure*}

Figure~\ref{fig:trajectories} shows the per-step HealthBench validation trajectories from which the best-by-val checkpoints in Section~\ref{sec:hb_new} are selected.

\section{Full cross-domain results}\label{app:cross_domain_full}

Tables~\ref{tab:cd_7b_new}, \ref{tab:cd_14b_new}, and~\ref{tab:cd_llama_new} report the complete four-task cross-domain suite (MMLU, GSM8K, IFEval, TruthfulQA) for all three base models, at each method's val-selected checkpoint. The body (Table~\ref{tab:cd_ifeval}) summarizes only the IFEval change, the single axis on which methods differ.

\begin{table}[t]
\centering
\caption{Cross-domain evaluation, Qwen2.5-7B-Instruct, at each method's val-selected checkpoint (matching Table~\ref{tab:hb_7b_new}). GSM8K under strict-match scoring. Fwd/Rev: forward/reverse KL; FT: full fine-tuning.}
\vspace{0.5em}
\scriptsize
\begin{tabular}{lllcccc}
\toprule
Method & KL & PI & MMLU & GSM8K & IFEval & TQA \\
\midrule
Base & --- & --- & 0.716 & 0.751 & 0.567 & 0.648 \\
GRPO & --- & rubric & 0.717 & 0.754 & 0.580 & 0.649 \\
\midrule
LoRA & Fwd & rubric & 0.716 & 0.800 & 0.521 & 0.642 \\
LoRA & Rev & rubric & 0.716 & 0.783 & 0.588 & 0.636 \\
FT & Fwd & rubric & 0.717 & 0.789 & 0.529 & 0.649 \\
FT & Rev & rubric & 0.716 & 0.757 & 0.560 & 0.638 \\
\midrule
LoRA & Fwd & reference & 0.716 & 0.782 & 0.558 & 0.645 \\
LoRA & Rev & reference & 0.715 & 0.744 & 0.586 & 0.647 \\
FT & Fwd & reference & 0.716 & 0.767 & 0.556 & 0.647 \\
FT & Rev & reference & 0.716 & 0.756 & 0.588 & 0.644 \\
\midrule
SFT & --- & \shortstack[l]{rubric\\(RCG)} & 0.718 & 0.715 & 0.488 & 0.613 \\
SFT & --- & reference & 0.720 & 0.748 & 0.429 & 0.623 \\
\bottomrule
\end{tabular}
\label{tab:cd_7b_new}
\end{table}

\begin{table}[t]
\centering
\caption{Cross-domain evaluation, Qwen2.5-14B-Instruct, at each method's val-selected checkpoint (matching Table~\ref{tab:hb_14b_new}). IFEval is prompt-level strict accuracy. The Qwen-14B base scores below the Qwen-7B base on IFEval (both strict, $0.518$ vs.\ $0.567$, and loose, $0.591$ vs.\ $0.638$) a genuine instruction-following inversion between these two instruct checkpoints, not a parsing artifact, so absolute IFEval is not comparable across scales; we interpret only the within-model change in IFEval (reported in the text), scoring base and trained checkpoints identically. GSM8K under strict-match scoring. Fwd/Rev: forward/reverse KL; FT: full fine-tuning. $^\dagger$RCG-SFT's answer formatting lowers strict-match GSM8K; flexible-extract recovers it to 0.785 $\approx$ base, so this is a format shift, not a reasoning loss.}
\vspace{0.5em}
\scriptsize
\begin{tabular}{lllcccc}
\toprule
Method & KL & PI & MMLU & GSM8K & IFEval & TQA \\
\midrule
Base & --- & --- & 0.789 & 0.787 & 0.518 & 0.691 \\
GRPO & --- & rubric & 0.788 & 0.781 & 0.514 & 0.692 \\
\midrule
LoRA & Fwd & rubric & 0.787 & 0.757 & 0.477 & 0.680 \\
LoRA & Rev & rubric & 0.788 & 0.762 & 0.486 & 0.687 \\
FT & Fwd & rubric & 0.787 & 0.773 & 0.484 & 0.688 \\
FT & Rev & rubric & 0.787 & 0.770 & 0.508 & 0.690 \\
\midrule
LoRA & Fwd & reference & 0.785 & 0.792 & 0.499 & 0.687 \\
LoRA & Rev & reference & 0.787 & 0.778 & 0.499 & 0.691 \\
FT & Fwd & reference & 0.788 & 0.789 & 0.512 & 0.688 \\
FT & Rev & reference & 0.788 & 0.779 & 0.506 & 0.687 \\
\midrule
SFT & --- & \shortstack[l]{rubric\\(RCG)} & 0.774 & 0.543$^\dagger$ & 0.366 & 0.604 \\
SFT & --- & reference & 0.786 & 0.660 & 0.370 & 0.632 \\
\bottomrule
\end{tabular}
\label{tab:cd_14b_new}
\end{table}

\begin{table}[t]
\centering
\caption{Cross-domain evaluation, Llama-3.1-8B-Instruct, at each method's val-selected checkpoint (matching Table~\ref{tab:hb_llama_new}). GSM8K under strict-match scoring. Fwd/Rev: forward/reverse KL; FT: full fine-tuning.}
\vspace{0.5em}
\scriptsize
\begin{tabular}{lllcccc}
\toprule
Method & KL & PI & MMLU & GSM8K & IFEval & TQA \\
\midrule
Base & --- & --- & 0.683 & 0.705 & 0.457 & 0.545 \\
GRPO & --- & rubric & 0.684 & 0.763 & 0.442 & 0.545 \\
\midrule
LoRA & Fwd & rubric & 0.683 & 0.703 & 0.471 & 0.567 \\
LoRA & Rev & rubric & 0.685 & 0.713 & 0.445 & 0.566 \\
FT & Fwd & rubric & 0.685 & 0.713 & 0.457 & 0.568 \\
FT & Rev & rubric & 0.687 & 0.699 & 0.447 & 0.563 \\
\midrule
LoRA & Fwd & reference & 0.685 & 0.705 & 0.451 & 0.561 \\
LoRA & Rev & reference & 0.684 & 0.701 & 0.462 & 0.559 \\
FT & Fwd & reference & 0.685 & 0.698 & 0.466 & 0.556 \\
FT & Rev & reference & 0.685 & 0.691 & 0.445 & 0.554 \\
\midrule
SFT & --- & \shortstack[l]{rubric\\(RCG)} & 0.684 & 0.740 & 0.462 & 0.563 \\
SFT & --- & reference & 0.680 & 0.752 & 0.427 & 0.568 \\
\bottomrule
\end{tabular}
\label{tab:cd_llama_new}
\end{table}

\section{SFT baselines on HealthBench Hard}\label{app:sft_hard}
Table~\ref{tab:hard_summary} compares the distillation and RL methods on the two
informative HealthBench Hard facets; here we report the same two facets for the
two SFT baselines, alongside Base and the strongest RuPI variant in Table~\ref{tab:sft_hard} (repeated from
Table~\ref{tab:hard_summary}) for reference. Consistent with the in-domain
results, RCG-SFT (which carries rubric information) exceeds reference-SFT on both
facets at all three scales. RuPI leads the cluster-level aggregate at every scale.
On the accuracy axis RuPI leads on Qwen~7B and Llama~8B; the one exception is the
Qwen-14B accuracy axis, where RCG-SFT's offline rubric-conditioned targets give it
an edge ($0.222$ vs.\ $0.156$). On Llama both SFT baselines register small
non-zero accuracy ($0.055$--$0.064$), below RuPI ($0.077$) but above the
reference-PI and GRPO distillation variants, which score zero (Table~\ref{tab:hard_summary}).

\begin{table}[h]
\centering
\caption{SFT baselines on HealthBench Hard: cluster-level aggregate (Cluster) and
accuracy axis (Acc), at each arm's val-selected checkpoint. ref-SFT:
reference-SFT; RCG-SFT: rubric-conditioned-generation SFT. Base and RuPI
(strongest variant) repeat Table~\ref{tab:hard_summary} for reference.
\textbf{Bold}: best in row among the four columns shown.}
\vspace{0.5em}
\footnotesize
\begin{tabular}{llcccc}
\toprule
Model & Metric & Base & RuPI & \shortstack[l]{ref-\\SFT} & \shortstack[l]{RCG-\\SFT} \\
\midrule
\multirow{2}{*}{Qwen 7B}  & Cluster & 0.503 & \textbf{0.542} & 0.464 & 0.528 \\
 & Acc & \textbf{0.043} & 0.038 & 0.033 & 0.036 \\
\midrule
\multirow{2}{*}{Qwen 14B} & Cluster & 0.698 & \textbf{0.778} & 0.677 & 0.696 \\
 & Acc & 0.106 & 0.156 & 0.112 & \textbf{0.222} \\
\midrule
\multirow{2}{*}{Llama 8B} & Cluster & 0.532 & \textbf{0.625} & 0.606 & 0.610 \\
 & Acc & 0.000 & \textbf{0.077} & 0.055 & 0.064 \\
\bottomrule
\end{tabular}
\label{tab:sft_hard}
\end{table}

\section{Effect of per-token KL clipping}\label{app:klclip}

The per-token KL clip ($\tau{=}5.0$) is the one hyperparameter where reverse KL is in principle most sensitive: the surrogate loss produces per-token weights with potentially heavy tails when student and teacher diverge, so a few outlier tokens could dominate the gradient. To check that clipping is necessary rather than cosmetic, we re-run the two reverse-KL rubric configurations on Qwen2.5-7B-Instruct with clipping effectively disabled ($\tau{=}10^{6}$, well above any per-token KL observed during training). Scores are HealthBench test ($n{=}2500$, GPT-4.1 judge, $T{=}1.0$) at the val-selected checkpoint.

\begin{table}[h]
\centering
\caption{Effect of per-token KL clipping on Qwen2.5-7B-Instruct (HealthBench test). Removing the clip moves the score by $\le 0.014$, well below the method-level effects.}
\vspace{0.5em}
\small
\begin{tabular}{lcc}
\toprule
Method (rubric PI, reverse KL) & Clip ($\tau{=}5$) & No clip \\
\midrule
LoRA & 0.203 & 0.199 \\
FT & 0.205 & 0.219 \\
\bottomrule
\end{tabular}
\label{tab:klclip_new}
\end{table}

Removing the clip changes the converged HealthBench test score by at most $0.014$: RuPI (LoRA) is essentially flat ($0.203 \to 0.199$) and RuPI (full fine-tuning) shifts only marginally ($0.205 \to 0.219$). The larger shift ($0.014$) exceeds the test standard error ($\le 0.008$), but is an order of magnitude smaller than the method-level differences we report ($0.03$--$0.08$). The clip threshold $\tau{=}5.0$ inherited from prior work does not appear to be necessary at this scale; we retain it in the main runs for consistency with prior recipes \citep{shenfeld2026self}.

\section{Full $\lambda$-ablation panels ($\lambda{=}0$ and $\lambda{=}0.5$)}\label{app:lambda_panel}

Table~\ref{tab:lambda0_full} reports all twelve off-policy ($\lambda{=}0$) runs on Qwen2.5-7B-Instruct underlying the contrast in Section~\ref{sec:onpolicy}: LoRA and full fine-tuning $\times$ rubric and reference PI $\times$ forward and reverse KL, with full fine-tuning additionally run under both a frozen cached teacher and an EMA teacher with regenerated targets. (The LoRA arm was run cached-only.) All values are HealthBench test scores ($n{=}2500$, GPT-4.1 judge, $T{=}1.0$) at the val-selected checkpoint; ``Cluster'' is the non-clipped cluster-level aggregate. The base model scores $0.169$ (cluster $0.635$). No configuration exceeds base, and within each PI type the cached and EMA teachers agree to within the test standard error ($\le 0.008$), confirming the on-/off-policy axis, not the teacher regime, drives the gap in Section~\ref{sec:onpolicy}.

\begin{table}[h]
\centering
\caption{Full $\lambda{=}0$ off-policy panel, Qwen2.5-7B-Instruct (HealthBench test, $n{=}2500$). Base $=0.169$ (cluster $0.635$).}
\vspace{0.5em}
\small
\begin{tabular}{lllcc}
\toprule
Recipe & PI & KL / Teacher & Test & Cluster \\
\midrule
LoRA & rubric    & fwd / cached & 0.154 & 0.633 \\
LoRA & rubric    & rev / cached & 0.150 & 0.608 \\
LoRA & reference & fwd / cached & 0.156 & 0.622 \\
LoRA & reference & rev / cached & 0.147 & 0.599 \\
\midrule
Full FT & rubric    & fwd / cached & 0.147 & 0.627 \\
Full FT & rubric    & fwd / EMA    & 0.140 & 0.616 \\
Full FT & rubric    & rev / cached & 0.153 & 0.624 \\
Full FT & rubric    & rev / EMA    & 0.143 & 0.603 \\
Full FT & reference & fwd / cached & 0.147 & 0.609 \\
Full FT & reference & fwd / EMA    & 0.146 & 0.608 \\
Full FT & reference & rev / cached & 0.146 & 0.611 \\
Full FT & reference & rev / EMA    & 0.147 & 0.608 \\
\bottomrule
\end{tabular}
\label{tab:lambda0_full}
\end{table}

Table~\ref{tab:lambda05_full} reports the $\lambda{=}0.5$ mixed panel on Qwen2.5-7B-Instruct (rubric PI only; reverse and forward KL; LoRA and full fine-tuning, the latter with cached and EMA teachers). Each step draws half its roll-outs on-policy from the student and half from the teacher. The best $\lambda{=}0.5$ rubric variant reaches $0.178$ test, above every $\lambda{=}0$ run ($\le 0.156$) and below the $\lambda{=}1$ best ($0.215$), placing the $\lambda{=}0.5$ midpoint on the monotonic on-policy trend of Section~\ref{sec:onpolicy}.

\begin{table}[h]
\centering
\caption{$\lambda{=}0.5$ mixed-rollout panel, Qwen2.5-7B-Instruct, rubric PI (HealthBench test, $n{=}2500$; val-selected checkpoint). Base $=0.169$ (cluster $0.635$).}
\vspace{0.5em}
\small
\begin{tabular}{lllcc}
\toprule
Recipe & PI & KL / Teacher & Test & Cluster \\
\midrule
LoRA & rubric & fwd / cached & 0.167 & 0.617 \\
LoRA & rubric & rev / cached & 0.177 & 0.623 \\
Full FT & rubric & fwd / cached & 0.161 & 0.631 \\
Full FT & rubric & fwd / EMA   & \textbf{0.178} & 0.637 \\
Full FT & rubric & rev / cached & 0.173 & 0.624 \\
\bottomrule
\end{tabular}
\label{tab:lambda05_full}
\end{table}

\section{RubricHub KL / injectability diagnostic}\label{app:rubrichub_kl}

We repeat the precondition test of Section~\ref{sec:diagnostic} on RubricHub Science \emph{before} training: does rubric conditioning shift the teacher's next-token distribution on the student's on-policy roll-outs more than reference conditioning? Using Qwen2.5-7B-Instruct as both teacher and student, we compute per-token teacher--student KL (Mean, and Body over the middle 80\% of each response) under rubric PI and best-of-six reference PI, on 100 held-out Science prompts, following the identical protocol of Section~\ref{sec:diagnostic}. As Table~\ref{tab:rubrichub_kl} shows, rubric PI produces $1.8\times$ more Body KL than reference PI ($0.328$ vs.\ $0.182$ nats), closely matching the HealthBench ratio ($1.7\times$) and its absolute magnitude ($0.326$ Body KL). The precondition for RuPI therefore holds on Science, consistent with the training results in Table~\ref{tab:rubrichub_7b}.

\begin{table}[h]
\centering
\caption{Per-token teacher-student KL on base Qwen2.5-7B-Instruct for RubricHub Science, scored on student on-policy roll-outs (rubric PI vs.\ best-of-six reference PI), over 100 held-out prompts. Body KL is the mean over the middle 80\% of each response.}
\vspace{0.5em}
\small
\begin{tabular}{lccc}
\toprule
PI type & Mean KL & Body KL \\
\midrule
Rubric PI            & $0.376$ & $0.328$ \\
Reference PI         & $0.199$ & $0.182$ \\
\midrule
Ratio (rubric/ref)   & $1.9\times$ & $1.8\times$ \\
\bottomrule
\end{tabular}
\label{tab:rubrichub_kl}
\end{table}

We also repeat the mass-concentration measurement of Section~\ref{sec:diagnostic} (Appendix~\ref{app:concentration}) on RubricHub Science: for each of 100 held-out prompts we take the best-of-six answer as the gold, sample $8$ candidate responses (GPT-4.1, $T{=}0.9$), keep the rubric-satisfying ones ($325$ candidates), and measure the base teacher's per-token NLL gap between these other-good responses and the gold under 
each PI (positive implies that the teacher assigns the gold lower NLL, i.e., higher probability than the other valid responses and the probability mass is concentrated on the gold). As Table~\ref{tab:rubrichub_conc} shows, under hard (reference) PI the gold is $0.84$ nats more probable per token than other valid responses, whereas under soft (rubric) PI the gap is $\approx 0$ ($+0.02$): the gold is thus $\exp(0.82)\approx 2.3\times$ more over-concentrated per token under hard PI. This mirrors the HealthBench mechanism ($6.5\times$): reference PI places probability mass on one arbitrary point in the valid set, while rubric PI spreads it across the set the student explores, which is why rubric PI yields the stronger per-token training signal in Table~\ref{tab:rubrichub_kl}.

\begin{table}[h]
\centering
\caption{Per-token mass concentration on base Qwen2.5-7B-Instruct for RubricHub Science (100 held-out prompts, 325 rubric-satisfying candidates). Over-concentration is the mean per-token NLL of other-good responses minus that of the gold; a larger positive value means the teacher concentrates more mass on the single gold answer.}
\vspace{0.5em}
\small
\begin{tabular}{lc}
\toprule
PI type & 
\shortstack[l]{Over-concentration\\(other-good $-$ gold, nats)}
\\
\midrule
Soft (rubric) PI    & $+0.02$ \\
Hard (reference) PI & $+0.84$ \\
\bottomrule
\end{tabular}
\label{tab:rubrichub_conc}
\end{table}

\section{GRPO baseline configuration}\label{app:grpo}
Our RaR GRPO baseline uses the following recipe: KL coefficient $\beta=0.1$, learning rate $5\mathrm{e}{-6}$, sampling temperature $1.2$, and $G=8$ completions per prompt over $32$ unique prompts per batch ($256$ generations per step), trained for $200$ steps. We set $\beta=0.1$ rather than the more common $\beta{=}0.04$ or $\beta{=}0$ because the un-penalized run was unstable: it reward-hacked the judge, converging to degenerate non-English output. The stabilized $\beta=0.1$ recipe was the strongest usable
GRPO baseline we obtained. The reward is the same GPT-4.1 rubric judge used for evaluation, so the baseline is not disadvantaged by any policy--judge mismatch; this is the identical recipe used for RubricHub (Section~\ref{sec:rubrichub}), differing only in the training corpus. We save a checkpoint every $50$ steps and select the one with the highest validation rubric-satisfaction (val split, $n{=}500$).

\end{document}